\documentclass[runningheads]{llncs}
\usepackage{multirow}
\usepackage{xcolor}
\usepackage{colortbl} % 引入此包以支持表格单元格底色 \cellcolor
\usepackage{pifont}
\usepackage{amssymb}
\usepackage{makecell}

\definecolor{fstcolor}{HTML}{FFB3B3} % 红色 (1st)
\definecolor{sndcolor}{HTML}{FFD9B3} % 橙色 (2nd)
\definecolor{trdcolor}{HTML}{FFFFB3} % 黄色 (3rd)

\definecolor{failgray}{HTML}{B0B0B0}
\definecolor{metagray}{HTML}{808080}
\newcommand{\oom}{\textcolor{failgray}{\textit{OOM}}}
\newcommand{\tl}{\textcolor{failgray}{\textit{TL}}}
\newcommand{\cg}[1]{\textcolor{metagray}{#1}}

\newcommand{\fst}[1]{\cellcolor{fstcolor}\textbf{#1}}
\newcommand{\snd}[1]{\cellcolor{sndcolor}#1}
\newcommand{\trd}[1]{\cellcolor{trdcolor}#1}

\newcommand{\best}[1]{\underline{\textbf{#1}}}
\usepackage{eccv}
\usepackage{eccvabbrv}
\usepackage{graphicx}
\usepackage{booktabs}
\usepackage[accsupp]{axessibility}
\usepackage{hyperref}
\usepackage{orcidlink}

\begin{document}
% ---------------------------------------------------------------
\title{GRF-Recon: Global Ray-Field Optimization for Long-Sequence Feed-forward Reconstruction} 
\titlerunning{GRF-Recon}
\author{Enpeng Li\orcidlink{0009-0006-8967-5662} \and
Yunzhou Zhang\orcidlink{0000-0002-2118-377X}\thanks{Corresponding author.} \and
Zhiyao Zhang \orcidlink{0009-0002-4864-7017} \and
Dexuan Lyu\orcidlink{0009-0004-7942-8823} \and
Chenyu Wang\orcidlink{0009-0005-6307-5522} \and
Chiyuan Cui\orcidlink{0009-0009-8570-7470} \and
Cheng Cheng\orcidlink{0009-0000-4027-0468}}

%\authorrunning{F.~Author et al.}
\authorrunning{E.~Li et al.}


\institute{College of Information Science and Engineering, Northeastern University, China \\
\email{zhangyunzhou@mail.neu.edu.cn}}
\maketitle

\begin{abstract}
Feed-forward 3D reconstruction provides an efficient paradigm for scene modeling from image sequences. Scaling these models to large monocular scenarios are constrained by excessive GPU memory footprint, degraded local geometry, and long-term trajectory drift. Existing chunk-based optimization strategies provide limited geometric constraints and fail to maintain global consistency over extended trajectories. We present a unified framework for stable and scalable feed-forward 3D reconstruction from long monocular sequences. Our approach builds on coarse-to-fine trajectory alignment augmented by lightweight geometric prior injection. Distilling monocular geometric cues into the feed-forward backbone via LoRA adaptation improves depth accuracy on fine structures while preserving inference efficiency. We introduce a hybrid-weight sparse ray-field optimization that leverages high-frequency geometric features to guide local point-cloud refinement and enforce consistent inter-frame ray constraints. Unlike prior chunk-based methods, this establishes strong cross-frame geometric coupling while maintaining scalability. Finally, an efficient trajectory stitching strategy with joint ray-error optimization explicitly reduces accumulated drift. Extensive experiments show that our approach achieves competitive trajectory accuracy compared with representative SLAM systems, while maintaining globally consistent 3D reconstruction in large-scale scenarios.
\end{abstract}

\begin{figure}[t]
  \centering
  \includegraphics[width=1.0\linewidth]{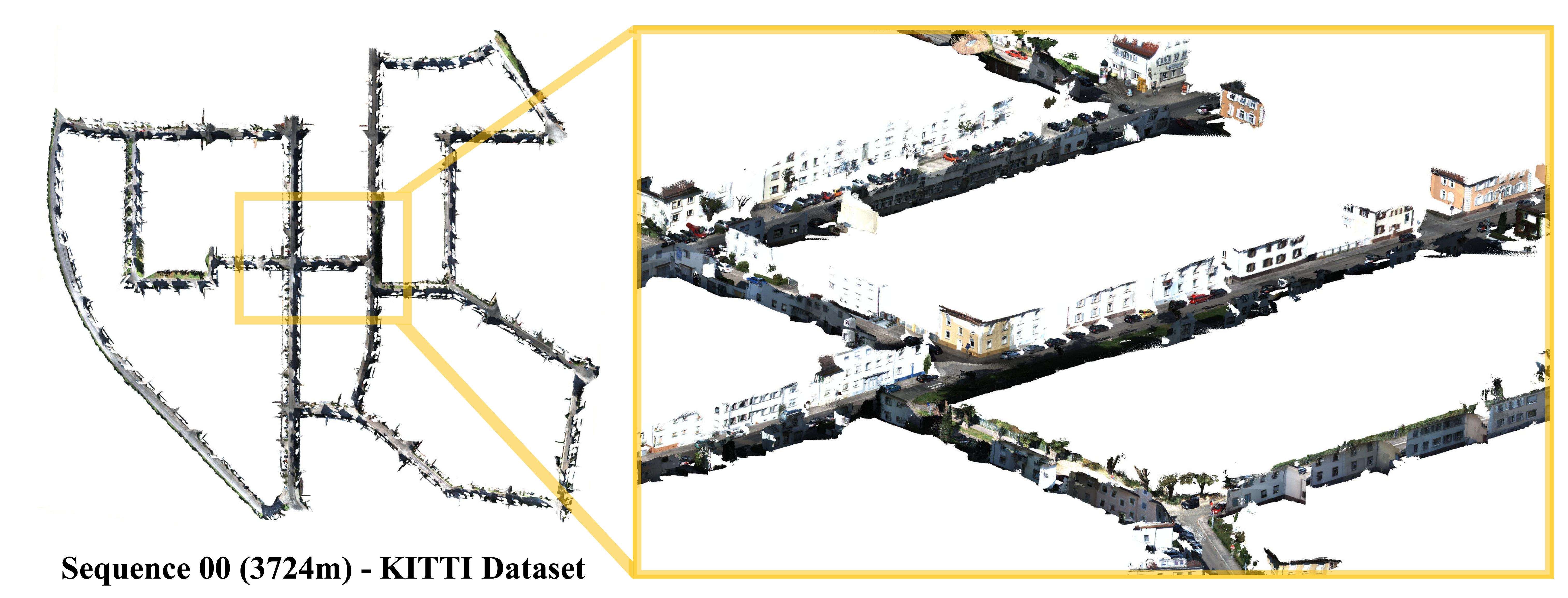} 
  \caption{\textbf{Large-scale 3D reconstruction using GRF-Recon.} Our proposed framework successfully reconstructs a globally consistent, low-drift point cloud map over a kilometer-scale monocular video sequence (KITTI~\cite{geiger2012we} Seq. 00, 3724m). The left panel shows the accurately closed global trajectory, while the right panel highlights the high-fidelity geometric details recovered at a complex urban intersection.}
  \label{fig:framework}
\end{figure}

\section{Introduction}
\label{sec:intro}
Achieving accurate camera pose estimation and dense 3D scene reconstruction from monocular RGB video streams remains a core challenge in 3D vision and autonomous driving~\cite{newcombe2011dtam, czarnowski2020deepfactors}. Due to the inherent inability of monocular vision to perceive absolute scale, the problem of accumulated drift in long-sequence tracking is particularly severe, often leading to significant global scale inconsistencies. Traditional systems for processing large-scale monocular sequences heavily rely on complex, multi-module pipelines consisting of camera calibration, visual odometry, multi-sensor fusion, and heavy backend graph optimization~\cite{campos2021orb, schonberger2016structure, triggs1999bundle}. In loosely coupled architectures, sensor noise and data association errors inevitably amplify and accumulate over large-scale movements.

Recently, Transformer-based~\cite{vaswani2017attention} end-to-end foundation models have emerged in 3D vision, enabling efficient pose inference and high-quality dense point cloud reconstruction. Ranging from DUSt3R~\cite{wang2024dust3r} and MASt3R~\cite{leroy2024grounding} to the latest VGGT~\cite{wang2025vggt} and Depth Anything 3 (DA3)~\cite{lin2026depth}, these models are pre-trained on massive datasets, integrating camera pose estimation, intrinsic regression, and 3D scene representation into a tightly coupled framework. Their core advantage lies in enabling the end-to-end backpropagation of errors throughout the entire system, thereby establishing a universal 3D reconstruction foundation model capable of directly processing uncalibrated, raw RGB images.

While these models perform well on benchmark datasets and short sequences, their scalability to complex, large-scale real-world scenarios remains limited. Their computational and memory requirements grow quadratically with the number of input tokens due to the Transformer attention mechanism, resulting in substantial GPU memory consumption during inference~\cite{rao2021dynamicvit, shen2026fastvggt}. In practice, this often restricts processing to only a few dozen frames on a consumer-grade GPU. Moreover, the lack of explicit global constraints makes long sequences susceptible to trajectory drift. Local estimation errors accumulate over time, leading to $\text{Sim}(3)$ drift that degrades multi-view consistency and metric accuracy~\cite{deng2026vggt, murai2025mast3r}. Robustness further deteriorates in large, unbounded scenes. Since these models are strongly influenced by their pre-training distributions, drastic viewpoint changes or unseen geometric configurations can destabilize local reconstruction, particularly around fine structures and depth discontinuities.

In long-sequence settings, scaling feed-forward foundation models requires reduced memory consumption and sensitivity to fine-grained geometric structures. We therefore incorporate feature redundancy filtering and lightweight geometric adaptation~\cite{ren2025finr, hu2022lora} to construct an efficient local perception front-end. Improving the local front-end alone does not address the challenge of maintaining global consistency over kilometer-scale trajectories, where loosely coupled chunk-stitching strategies often fail to enforce long-range geometric constraints. To overcome this limitation, we introduce \textbf{GRF-Recon}, a monocular reconstruction system for large-scale scenes (Fig.~\ref{fig:framework}). It formulates a global joint optimization scheme based on a hybrid-weight sparse ray-field to explicitly couple cross-view geometric constraints. Experiments on large-scale outdoor autonomous driving benchmarks, including KITTI~\cite{geiger2012we}, show that our method achieves accurate reconstruction of kilometer-scale environments without requiring camera calibration or depth supervision. It outperforms prior feed-forward models and achieves competitive performance compared to recent SLAM-based approaches in trajectory accuracy and global map consistency~\cite{teed2021droid, zhang2023go}.

\section{Related Work}
\textbf{Monocular 3D Reconstruction and SLAM.} Classical Structure-from-Motion (SfM) and visual SLAM systems estimate camera poses and sparse 3D structures either implicitly or explicitly via multi-view geometry~\cite{hartley2003multiple}. Traditional pipelines like COLMAP~\cite{schonberger2016structure} typically follow an incremental paradigm, relying heavily on feature point detection~\cite{detone2018superpoint, tyszkiewicz2020disk}, cross-view matching~\cite{sarlin2020superglue, lindenberger2023lightglue}, and heavy bundle adjustment~\cite{triggs1999bundle}. While these methods demonstrate high robustness in ideal environments, they exhibit extreme vulnerability in textureless regions and accumulate irreversible scale drift over long sequences. Recently, fully differentiable systems like VGGSfM~\cite{wang2024vggsfm} have demonstrated the immense potential of end-to-end learning frameworks through pixel-level 2D point tracking and global simultaneous recovery mechanisms. In terms of dense matching and tracking, DROID-SLAM~\cite{teed2021droid} innovatively integrates deep learning features with dense bundle adjustment. Its backend essentially inherits the optimization logic of sparse SLAM. Lacking explicit global geometric constraints, it remains prone to 3D geometric inconsistencies in long sequences~\cite{lipson2024deep}. Recent efforts such as MASt3R-SLAM~\cite{murai2025mast3r} attempt to embed feed-forward priors into real-time SLAM but still require complex traditional backend machinery.

\textbf{Feed-forward 3D Foundation Models and Long-Sequence Scaling.} The most significant recent paradigm shift in 3D vision is the emergence of end-to-end foundation models based on the Transformer architecture~\cite{vaswani2017attention, wang2024dust3r, leroy2024grounding}. These methods are capable of directly generating remarkably stable local 3D maps from uncalibrated overlapping images. The immense computational and memory overhead induced by the self-attention mechanism strictly limits their application to relatively short sequences. To break the bottleneck of processing long sequences, Spann3R~\cite{wang20253dspann3r} and LONG3R~\cite{chen2025long3r} explore streaming reconstruction based on spatio-temporal memory, whereas MUSt3R~\cite{cabon2025must3r} and Fast3R~\cite{yang2025fast3r} focus on multi-view parallelization to eliminate pairwise matching errors. Despite some progress, streaming processing is susceptible to long-term feature drift, and parallel networks still hit physical memory ceilings when facing ultra-long trajectories. VGGT-Long~\cite{deng2026vggt} explores a minimalist "chunk-and-align" strategy. While the chunking paradigm initially alleviates the memory crisis, existing chunk stitching methods lack deep geometric constraints~\cite{wang2026pi}.

\textbf{Efficient 3D Perception and Feature Redundancy Filtering.} As feed-forward foundation models process increasing viewpoints and resolutions, the proliferation of tokens in Transformers leads to high inference costs. In real-world 3D reconstruction, the density of geometric information across different spatial regions is highly uneven, with substantial textureless backgrounds and overlapping viewpoint redundancies. Recently, dynamic token pruning~\cite{rao2021dynamicvit} and token merging~\cite{bolya2022token} techniques in visual architectures have been studied to accelerate inference. In the realm of 3D foundation models, Fast-VGGT~\cite{shen2026fastvggt} proposes an efficient feature filtering paradigm. By identifying and suppressing redundant features in overlapping views and stable regions, the model significantly reduces computational burden and memory consumption with almost no loss in local reconstruction accuracy. To recover high-frequency geometric details while obtaining accurate poses, the research community (e.g., Fin3R~\cite{ren2025finr}) has introduced Parameter-Efficient Fine-Tuning and monocular knowledge distillation techniques to enhance the model's perceptual acuity~\cite{birkl2023midas, wang2025moge}.

In this work, we propose a hybrid-weight sparse ray-field that constructs rigorous explicit geometric constraints. We leverage LoRA-based~\cite{hu2022lora} lightweight prior injection to accurately restore local geometric edges and ultimately establish strict global physical constraints. With minimal system overhead, this framework successfully and stably extends the boundaries of feed-forward monocular 3D reconstruction to kilometer-scale unbounded scenes.

\section{Method}
\subsection{Local Depth Refinement and Initial Trajectory Stitching}
\label{subsec:local_refinement}

\begin{figure}[t]
  \centering
  \includegraphics[width=1.0\linewidth]{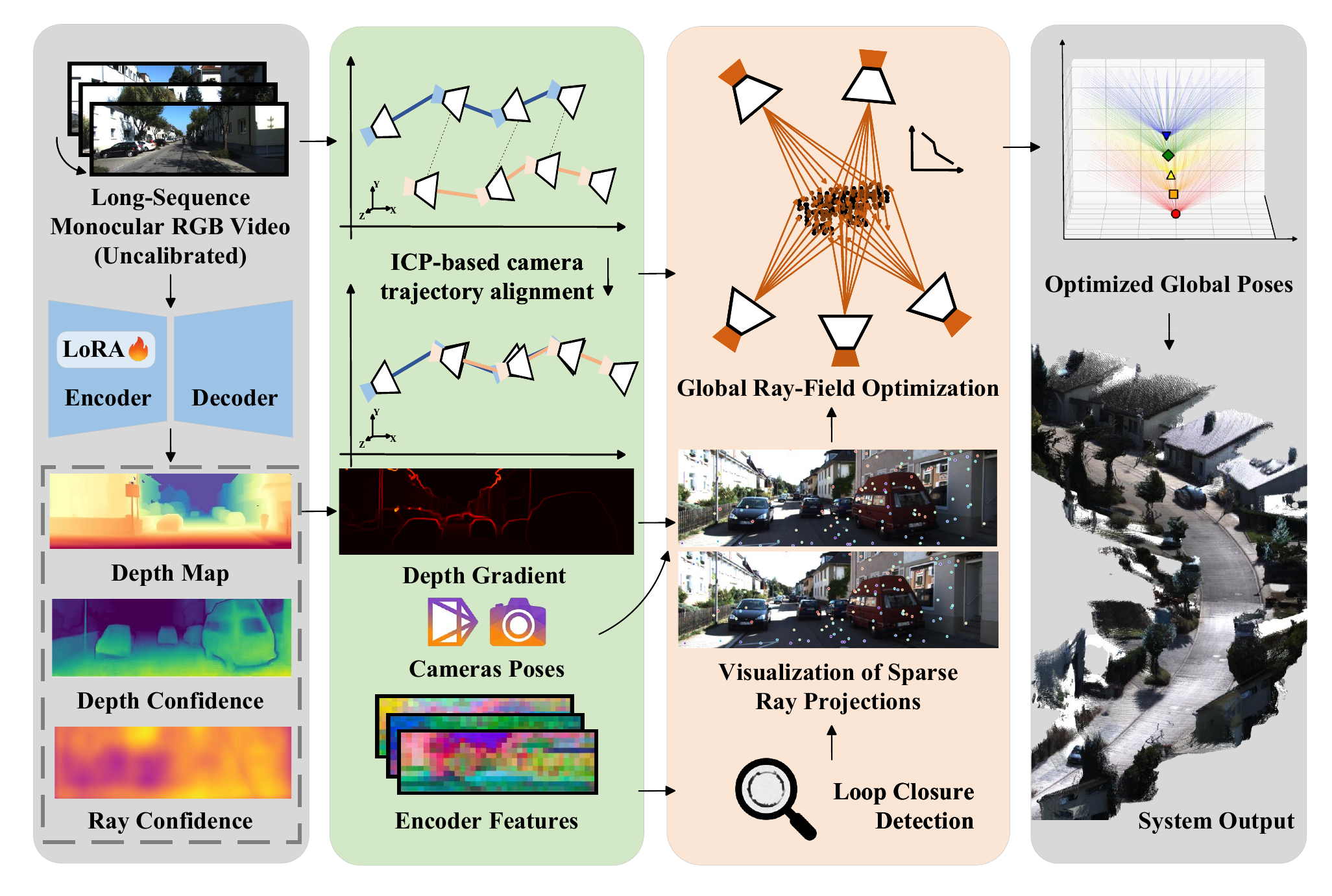} 
   \caption{\textbf{Overview of the GRF-Recon System Framework.} GRF-Recon reconstructs 3D geometry from uncalibrated long-sequence monocular RGB videos, building on a pretrained model with LoRA fine-tuning to enhance geometric perception. The system adopts a coarse-to-fine chunk-based parallel pipeline, performing confidence-guided sparse ray sampling and hybrid-weighted ray matching to enforce cross-view consistency. The orange module visualizes the projection of ray maps from neighboring frames onto the 2D image plane. Global optimization is carried out in a factor graph that jointly incorporates ray-field and loop-closure constraints, producing refined camera trajectories and a globally consistent, high-fidelity 3D point cloud.}
  \label{fig:OS}
\end{figure}

The proposed GRF-Recon system, an overview of which is illustrated in Fig.~\ref{fig:OS}, utilizes DA3~\cite{lin2026depth} as the perception backbone, introducing enhancements to enable low-drift 3D reconstruction over long sequences. Although DA3 is pre-trained on large-scale datasets, applying it to sequential reconstruction often suffers from geometric over-smoothing and degradation near occlusion boundaries. Inspired by Fin3R~\cite{ren2025finr}, we adopt a re-normalized Low-Rank Adaptation~\cite{hu2022lora} strategy to fine-tune the DA3 encoder. This operation injects high-frequency monocular geometric priors while preserving the model's multi-view matching generalization ability, thereby improving local reconstruction fidelity.

To implement this fine-tuning without degrading DA3's cross-view correlation capability, we freeze the decoder and insert low-rank adapters into the Query–Key–Value projections of the DINOv2~\cite{oquab2024dinov2} encoder attention layers. However, naive monocular distillation causes a feature norm drift, leading to a mismatch with the frozen decoder's expected input distribution. We address this by constraining the updated effective weight $\mathbf{W}'$ via a re-normalization term:

\begin{equation}
    \mathbf{W}' = \frac{(\mathbf{W}_0 + \mathbf{B}\mathbf{A}) \cdot \|\mathbf{W}_0\|_F}{\|\mathbf{W}_0 + \mathbf{B}\mathbf{A}\|_F}
    \label{eq:lora_norm}
\end{equation}
where $\|\cdot\|_F$ denotes the Frobenius norm. This strategy introduces less than 1\% additional trainable parameters while preserving the original weight norm, effectively stabilizing the feature activation distribution.

Following the encoder adaptation, we mitigate the risk of dataset overfitting during direct supervision by employing a teacher–student dual-branch distillation strategy. Specifically, a depth branch leverages high-quality pseudo-labels generated by an advanced monocular geometry estimator~\cite{wang2025moge}, applying a log-uncertainty-weighted $\ell_2$ loss to supervise the DA3 depth predictions. Concurrently, a point branch utilizes ground-truth 3D point clouds to provide regression supervision, ensuring fundamental multi-view geometric consistency. The overall distillation loss is thus defined as $\mathcal{L} = \mathcal{L}_{\text{depth}} + \lambda \mathcal{L}_{\text{point}}$, where $\lambda = 0.4$ balances the two modalities.

Once the perception front-end is capable of producing high-fidelity depth and ray-field, we process the long sequences using a chunk-based alignment strategy. To merge isolated local chunks into a unified global coordinate system, we extract the poses of the overlapping frames from two neighboring chunks $\mathcal{C}_k$ and $\mathcal{C}_{k+1}$. We compute the similarity transformation $\mathbf{M}_{\text{sim}} \in \mathrm{Sim}(3)$ via a decoupled optimization process. The rotation covariance matrix $\mathbf{Q}$ is formulated from the overlapping frames, and the optimal rotation $\mathbf{R}^*$ is derived through Singular Value Decomposition (SVD):
\begin{equation}
    \mathbf{U}, \boldsymbol{\Sigma}, \mathbf{V}^\top = \mathrm{SVD}(\mathbf{Q}), \quad \mathbf{R}^* = \mathbf{U} \, \mathrm{diag}(1, 1, \det(\mathbf{U}\mathbf{V}^\top)) \, \mathbf{V}^\top
    \label{eq:svd_r}
\end{equation}
We solve for the scale $s^*$ and translation $\mathbf{t}^*$ via a least-squares formulation:
\begin{equation}
    s^*, \mathbf{t}^* = \mathop{\arg\min}\limits_{s, \mathbf{t}} \sum_{i=1}^{m} \left\| \mathbf{t}_k^{(i)} - \left( s \mathbf{R}^* \mathbf{t}_{k+1}^{(i)} + \mathbf{t} \right) \right\|_2^2
    \label{eq:least_squares}
\end{equation}
The resulting matrix $\mathbf{M}_{\text{sim}}$ is applied to align $\mathcal{C}_{k+1}$ with $\mathcal{C}_k$. By propagating this spatial alignment along the sequence, the system establishes a coarse but globally consistent initial trajectory $\mathcal{T}_{\text{global}}$.

\subsection{Hybrid Sparse Ray-Field Construction}
\label{subsec:hybrid_ray_field}

Unlike traditional methods that strictly rely on pinhole models for initial geometric reasoning, our perceptual front-end directly utilizes the DA3 ray decoding head to predict pixel-level features end-to-end. For a given pixel $p_i$ in the source frame $I_i$, the network outputs a 7-channel ray vector. The first 6 channels formulate the 3D ray parameterized as $\mathbf{r}_i = [\mathbf{d}_i^\top, \mathbf{o}_i^\top]^\top \in \mathbb{R}^6$, where $\mathbf{d}_i$ is the normalized ray direction and $\mathbf{o}_i$ indicates the ray origin in the local coordinate system. The 7th channel yields the ray prediction confidence $c^{\text{ray}}_{i}$. Simultaneously, the depth branch outputs the metric depth value $z_i$ along with its associated confidence $c^{\text{depth}}_{i}$. Leveraging these metric properties, the local 3D point corresponding to the pixel is precisely reconstructed as $\mathbf{P}_i = \mathbf{o}_i + z_i \mathbf{d}_i$.

To manage the matching complexity inherent in long sequences, we sample the predicted rays sparsely based on a dual confidence fusion mechanism:
\begin{equation}
    c^{\text{joint}}_{i} = \beta \cdot \hat{c}^{\text{depth}}_{i} + (1-\beta) \cdot \hat{c}^{\text{ray}}_{i}
    \label{eq:joint_conf}
\end{equation}
where $\beta=0.5$ acts as a trade-off parameter, and $\hat{c}^{\text{depth}}_{i}$, $\hat{c}^{\text{ray}}_{i}$ are the min-max normalized confidences. We then incorporate the depth gradient magnitude $\hat{g}_{i} = \|\nabla D(i)\|_2$ to construct a composite score $s_{i} = \gamma_{\text{conf}} \cdot c^{\text{joint}}_{i} + \gamma_{\text{grad}} \cdot \hat{g}_{i}$ ($\gamma_{\text{conf}}=0.8$ and $\gamma_{\text{grad}}=0.2$). Pixels satisfying a predefined confidence threshold are sampled according to the probability distribution $\mathcal{P}(i) \propto s_i$, followed by Non-Maximum Suppression to ensure a uniform spatial distribution across the image.

We establish multidimensional local constraints to associate observations across frames. For a reconstructed 3D point $\mathbf{P}_i$ originating from $p_i$, we determine its initial search anchor in the target frame $I_j$ via perspective projection: $\mathbf{p}_j^{\text{init}} = \pi(\mathbf{K}_j (\mathbf{R}_j \mathbf{P}_i + \mathbf{t}_j))$, where $\pi(\cdot)$ denotes homogeneous normalization and $\mathbf{K}_j$ represents the camera intrinsics. Using this anchor as the center, feature matching is performed within a local search window $\mathcal{W}$. The primary geometric criterion is the ray distance metric $e_{\text{ray}}$, defined as the perpendicular Euclidean distance from the target candidate ray $\mathbf{r}_j$ to the point $\mathbf{P}_i$:
\begin{equation}
    e_{\text{ray}}(\mathbf{P}_i, \mathbf{r}_j) = \frac{\|(\mathbf{P}_i - \mathbf{o}_j) \times \mathbf{d}_j\|_2}{\|\mathbf{d}_j\|_2}
    \label{eq:ray_dist}
\end{equation}
To further disambiguate structural and textural similarities, we introduce a gradient similarity term $e_{\text{grad}} = 1 - (\nabla I_i^\top \nabla I_j) / (\|\nabla I_i\| \|\nabla I_j\|)$ and a photometric error $e_{\text{photo}}$, computed via the Sum of Absolute Differences (SAD) over a local patch. The comprehensive matching score is thus constructed as:
\begin{equation}
    \mathcal{S}(p_i, p_j) = w_r \cdot e_{\text{ray}} + w_g \cdot e_{\text{grad}} + w_p \cdot e_{\text{photo}}
    \label{eq:total_score}
\end{equation}

where $w_r = 0.6$, $w_g = 0.2$, and $w_p = 0.2$ are the optimal weighting coefficients (evaluated in Sec.~\ref{subsec:ablation}). The target match $p_j^*$ is obtained by minimizing this composite score $\mathcal{S}$. To filter out spurious correspondences, bidirectional symmetry verification is conducted: the 3D point reconstructed from $p_j^*$ is projected back to the source frame. The match is accepted only if the reprojection error satisfies $\sqrt{(u_i^{\text{back}} - u_i)^2 + (v_i^{\text{back}} - v_i)^2} < \tau_{\text{sym}}$, with $\tau_{\text{sym}} = 5.0$.

\subsection{Loop Detection and Global Graph Optimization}
\label{subsec:global_optimization}

To balance geometric constraints and computational overhead, we dynamically extract keyframes and detect loop closures using a unified feature similarity metric. For any two frames $I_t$ and $I_s$, we extract DINOv2 global descriptors $f_t$ and $f_s$, and compute their visual cosine similarity~\cite{tolias2013aggregate}:
\begin{equation}
    \mathrm{sim}(I_t, I_s) = \frac{f_t^\top f_s}{\|f_t\| \|f_s\|}
    \label{eq:dino_sim}
\end{equation}

Based on this metric, a new keyframe is triggered when the similarity between the current frame and the last keyframe drops below a threshold $\tau_{kf} = 0.70$. This adaptive strategy ensures dense sampling during complex viewpoint transitions and sparse sampling during smooth linear motion, empirically yielding an average interval of 3 to 5 frames on the KITTI~\cite{geiger2012we} sequences. To recognize revisited areas and close the trajectory loops, a valid loop candidate is identified when the similarity between the current keyframe and a historical keyframe exceeds a strict threshold $\tau_{\text{sim}}$ (set to $0.85$), provided that the temporal interval ($\Delta t > 100$ frames) is sufficient to avoid redundant local constraints.

Upon detecting a loop candidate, we expand the temporal windows to include adjacent frames and leverage DA3 to extract high-confidence relative pose transformations $\hat{\mathbf{T}}_{ij}$.  To eliminate the accumulated drift and achieve global consistency, we jointly optimize the entire set of camera poses $\{\mathbf{T}_k \in SE(3)\}$~\cite{sola2021micro} using a unified factor graph. The global objective function balances the hybrid ray-field constraints against the multi-frame loop pose priors:
\begin{equation}
    E(\{\mathbf{T}_k\}) = \sum_{i,j,p} \rho \left( e_{\text{ray}}(\mathbf{T}_j^{-1} \mathbf{T}_i \mathbf{P}_i, \mathbf{r}_j) \right) + \lambda_{\text{pose}} \sum_{i,j} \rho_{\text{pose}} ( \mathbf{T}_i, \mathbf{T}_j, \hat{\mathbf{T}}_{ij} )
    \label{eq:global_energy}
\end{equation}
where the first term ensures geometric consistency across sparse ray-fields by transforming local points $\mathbf{P}_i$ into the target frame $j$, and the second term anchors the trajectory using the multi-frame predictions, weighted by $\lambda_{\text{pose}}$ (set to 10.0 to balance the point-level scale). The function $\rho(\cdot)$ denotes the robust Huber loss. With the metric-scale initialization provided by the piece-wise trajectory alignment (Sec.~\ref{subsec:local_refinement}), the Levenberg-Marquardt algorithm converges, yielding a globally consistent reconstruction with reduced drift.

\section{Experiments}
\label{sec:Experiments}
We evaluate GRF-Recon on large-scale, long-sequence monocular 3D reconstruction using outdoor driving benchmarks and standard depth datasets. Trajectory estimation and scene reconstruction are evaluated on KITTI~\cite{geiger2012we} and the Waymo Open Dataset~\cite{sun2020scalability}, while depth estimation is evaluated on ETH3D~\cite{schops2017multi}, SINTEL~\cite{baker2011database}, and DIODE~\cite{vasiljevic2019diode}. All experiments are conducted on a local workstation running Ubuntu 22.04, equipped with an Intel Core i9-13900K CPU, 128 GB RAM, and a single NVIDIA GeForce RTX 3090 GPU (24 GB).

\subsection{Experimental Setup}

\textbf{Evaluation Metrics.}
We adopt standard protocols to assess performance across all tasks. For long-sequence camera tracking, we report the absolute trajectory error as RMSE (ATE RMSE [m] $\downarrow$) to measure global consistency and accumulated drift. For monocular depth estimation, we evaluate relative error (Rel $\downarrow$) and threshold accuracy ($\delta_1 \uparrow$), capturing both metric fidelity and scale stability. For large-scale point cloud reconstruction, we measure point-level geometric accuracy using Accuracy ($\downarrow$), Completeness ($\downarrow$), and Chamfer Distance ($\downarrow$). Given the scale ambiguity inherent in monocular predictions, all estimated trajectories and point clouds are aligned to LiDAR ground truth via a global Sim(3) transformation before metric computation. To assess downstream novel-view synthesis quality, we report PSNR and SSIM to measure rendering fidelity.

\textbf{Data Preprocessing and Ground Truth.}
During inference, we resize each input image to a width of 504 pixels while preserving the aspect ratio, rounding the other dimension to a multiple of 14 to match the patch partitioning of the ViT-based front end. To ensure fair comparison, we apply a unified point filtering rule: for methods predicting confidence, we retain 3D points with a confidence score greater than 0.7 times the global mean. 

In autonomous driving datasets like the Waymo Open Dataset~\cite{sun2020scalability}, the vehicle-mounted LiDAR ground truth is physically constrained by sensor height and vertical field of view, leaving upper structures such as building tops and tree canopies unobserved. Vision-based feed-forward reconstruction successfully recovers these regions. When evaluated against sparse LiDAR references, these valid geometric reconstructions are often heavily penalized under distance-based metrics such as Accuracy (Fig.~\ref{fig:local_detail}). Consequently, qualitative visual comparison serves as an essential complement to quantitative evaluation when assessing the true geometric reconstruction capabilities.

\begin{figure}[t]
  \centering
  \includegraphics[width=1.0\linewidth]{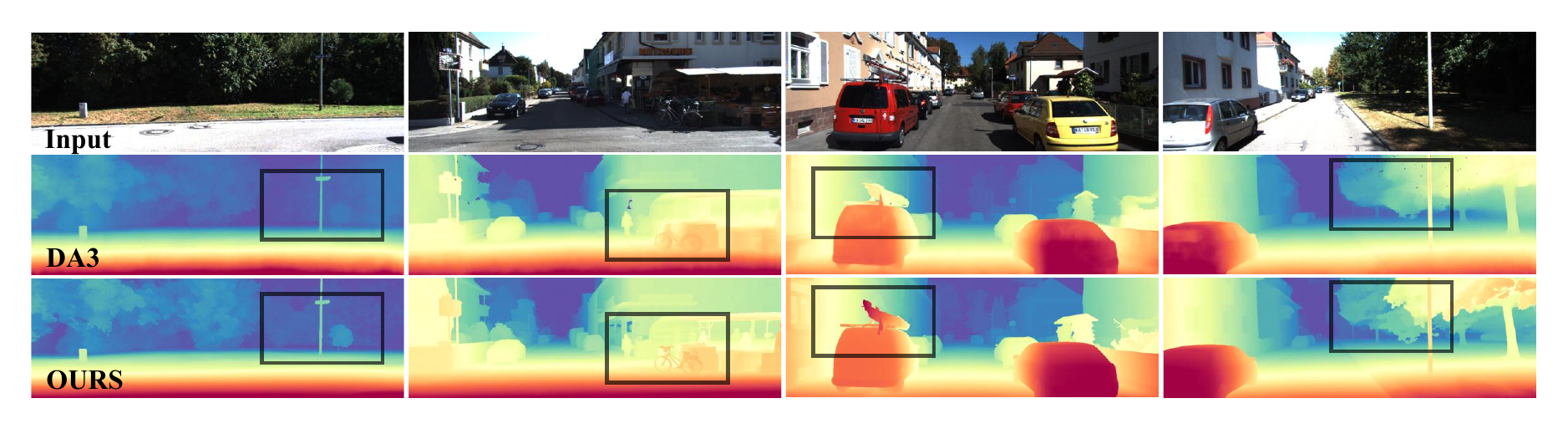} 
  \caption{\textbf{Qualitative comparison of monocular depth estimation.} We compare the baseline DA3 with the LoRA-enhanced model (Ours) on the KITTI dataset. The original DA3, using the DA3 NESTED-GIANT-LARGE-1.1 weights, exhibits over-smoothing and fails to capture thin, high-frequency structures. We fine-tune this model, and experiments show that it can recover sharp geometric details and cleanly separate slender utility poles, foliage, and complex vehicle structures from the background.}
  \label{fig:framework_depth}
\end{figure}

\subsection{Monocular Depth Estimation}

High-fidelity local geometric priors underpin the hybrid-weight ray-field and enable accurate correspondence search. Since LiDAR scans in the original KITTI odometry split are highly sparse, we evaluate depth on the semi-dense ground truth provided by the KITTI Depth~\cite{geiger2012we} (Fig.~\ref{fig:framework_depth}). Zero-shot generalization across diverse scene geometries is assessed on ETH3D~\cite{schops2017multi}, SINTEL~\cite{baker2011database}, and DIODE~\cite{vasiljevic2019diode}.

The perception backbone is initialized from a pre-trained DA3~\cite{lin2026depth} model. Following a monocular knowledge-distillation paradigm, the decoder responsible for cross-view feature matching remains frozen. Only the image encoder and depth head are updated. To prevent feature distribution shift, we insert a re-normalized low-rank adaptation (Re-normalized LoRA, $r=8$) layer into the encoder. Fine-tuning runs on a single NVIDIA RTX 3090 with automatic mixed precision (AMP), using a batch size of 16 for 15 epochs ($\approx 14$ hours). The total parameter increase remains strictly below 3\%.

Tab.~\ref{tab:depth_estimation_updated} summarizes the quantitative results. After fine-tuning (DA3+Ours), front-end depth accuracy improves significantly: on KITTI, Rel decreases from 5.14 to 4.50, and $\delta_1$ increases from 92.4 to 97.1. On ETH3D and dynamic high-illumination scenes, the model maintains high stability, reducing average Rel to 6.95 and pushing average $\delta_1$ to 91.4, tightly approaching the teacher model (MoGe~\cite{wang2025moge}). Compared to full fine-tuning, our LoRA-based scheme effectively mitigates overfitting while producing sharp, high-fidelity local geometry. These precise priors constrain the search space for subsequent ray-field optimization, laying a solid foundation for stable long-sequence reconstruction.

\begin{table}[t]
\centering
\scriptsize
\setlength{\tabcolsep}{2pt}
\renewcommand{\arraystretch}{0.95}
\caption{Quantitative results for monocular depth estimation.}
\label{tab:depth_estimation_updated}
\begin{tabular}{l cc cc cc cc cc}
\toprule
\multirow{2}{*}{\textbf{Method}} &
\multicolumn{2}{c}{\textbf{KITTI}~\cite{geiger2012we}} &
\multicolumn{2}{c}{\textbf{ETH3D}~\cite{schops2017multi}} &
\multicolumn{2}{c}{\textbf{SINTEL}~\cite{baker2011database}} &
\multicolumn{2}{c}{\textbf{DIODE}~\cite{vasiljevic2019diode}} &
\multicolumn{2}{c}{\textbf{Avg}} \\
\cmidrule(lr){2-3} \cmidrule(lr){4-5}
\cmidrule(lr){6-7} \cmidrule(lr){8-9}
\cmidrule(l){10-11}
& Rel $\downarrow$ & $\delta_1 \uparrow$
& Rel $\downarrow$ & $\delta_1 \uparrow$
& Rel $\downarrow$ & $\delta_1 \uparrow$
& Rel $\downarrow$ & $\delta_1 \uparrow$
& Rel $\downarrow$ & $\delta_1 \uparrow$ \\
\midrule
DA3~\cite{lin2026depth} & 5.14 & 92.4 & 3.25 & 97.6 & 16.20 & 73.5 & 6.31 & 94.2 & 7.73 & 89.4 \\
DA3+Ours & \underline{4.50} & \underline{97.1}
& \underline{2.48} & \underline{99.2}
& \underline{15.21} & \underline{74.3}
& \underline{5.62} & \underline{94.9}
& \underline{6.95} & \underline{91.4} \\
\midrule
\rowcolor{gray!10}
MoGe (Teacher)~\cite{wang2025moge} & \textbf{4.39} & \textbf{97.2}
& \textbf{2.10} & \textbf{99.8}
& \textbf{11.50} & \textbf{81.4}
& \textbf{4.50} & \textbf{96.6}
& \textbf{5.62} & \textbf{93.8} \\
\bottomrule
\end{tabular}
\end{table}

\begin{table*}[t]
    \centering
    \caption{\textbf{Camera Tracking on KITTI}~\cite{geiger2012we} (ATE RMSE [m] $\downarrow$). To ensure a fair comparison, the Average (Avg.) is computed by excluding the high-speed Seq. 01, which typically causes catastrophic scale drift in monocular methods. Methods are categorized based on their requirement for camera intrinsic calibration. \textit{TL} and \textit{OOM} denote Tracking Lost and Out-of-Memory, respectively. The top three results are highlighted in \fst{red}, \snd{orange}, and \trd{yellow}.}
    \label{tab:kitti_tracking_heatmap_updated}
    \resizebox{\textwidth}{!}{
    \begin{tabular}{c l | c | ccccccccccc}
        \toprule
        \multicolumn{2}{c|}{\textbf{Method}} & \textbf{Avg.} & \textbf{00} & \textbf{01} & \textbf{02} & \textbf{03} & \textbf{04} & \textbf{05} & \textbf{06} & \textbf{07} & \textbf{08} & \textbf{09} & \textbf{10} \\
        \midrule
        
        % Metadata 
        \multirow{4}{*}{\rotatebox{90}{\cg{\textit{Metadata}}}} 
        & \cg{\textit{seq. frames}} & \cg{2210} & \cg{4542} & \cg{1101} & \cg{4661} & \cg{801} & \cg{271} & \cg{2761} & \cg{1201} & \cg{1101} & \cg{4071} & \cg{1591} & \cg{1201} \\
        & \cg{\textit{seq. length (m)}} & \cg{1968.15} & \cg{3724.19} & \cg{2453.20} & \cg{5067.23} & \cg{560.89} & \cg{393.65} & \cg{2205.58} & \cg{1232.88} & \cg{649.70} & \cg{3222.80} & \cg{1705.05} & \cg{919.52} \\
        & \cg{\textit{seq. speed (m/f)}} & \cg{0.89} & \cg{0.82} & \cg{2.23} & \cg{1.09} & \cg{0.70} & \cg{1.45} & \cg{0.80} & \cg{1.12} & \cg{0.59} & \cg{0.79} & \cg{1.07} & \cg{0.77} \\
        & \cg{\textit{contains loop}} & \cg{-} & \cg{$\checkmark$} & \cg{$\times$} & \cg{$\checkmark$} & \cg{$\times$} & \cg{$\times$} & \cg{$\checkmark$} & \cg{$\checkmark$} & \cg{$\checkmark$} & \cg{$\times$} & \cg{$\checkmark$} & \cg{$\times$} \\
        
        \midrule
        
        \multirow{5}{*}{\rotatebox{90}{\textbf{Calibrated}}} 
        & ORB-SLAM3~\cite{campos2021orb} & \snd{7.93} & \snd{5.12} & \fst{7.23} & \snd{11.50} & \fst{1.35} & 2.15 & \snd{4.60} & 13.41 & \fst{1.55} & \fst{25.30} & \snd{8.50} & \fst{5.85} \\
        & DROID-SLAM~\cite{teed2021droid} & 74.88 & 95.10 & 334.20 & 117.31 & 4.38 & 2.20 & 128.50 & 54.47 & 15.58 & 151.60 & 67.33 & 112.32 \\
        & DPVO~\cite{teed2023deep} & 54.79 & 123.34 & \snd{8.29} & 113.34 & 4.09 & 1.68 & 48.56 & 44.78 & 20.36 & 104.21 & 64.10 & 23.43 \\
        & DPV-SLAM~\cite{lipson2024deep} & 59.04 & 113.20 & \trd{9.50} & 132.53 & \trd{2.31} & \fst{0.79} & 54.80 & 52.34 & 16.37 & 132.44 & 74.26 & 11.35 \\
        & DPV-SLAM++~\cite{lipson2024deep} & 24.25 & \trd{7.67} & 10.36 & \fst{10.36} & 3.45 & 1.28 & \trd{5.34} & 13.57 & \trd{2.01} & 124.46 & 63.64 & \trd{10.67} \\
        \midrule
        
        \multirow{7}{*}{\rotatebox{90}{\textbf{Uncalibrated}}} 
        & MASt3R-SLAM~\cite{murai2025mast3r} & \cg{-} & \tl & \tl & \tl & \tl & \tl & \tl & \tl & \tl & \tl & \tl & \tl \\
        & VGGT~\cite{wang2025vggt} & \cg{-} & \oom & \oom & \oom & \oom & \oom & \oom & \oom & \oom & \oom & \oom & \oom \\
        & DA3~\cite{lin2026depth} & \cg{-} & \oom & \oom & \oom & \oom & \oom & \oom & \oom & \oom & \oom & \oom & \oom \\
        & VGGT-Long~\cite{deng2026vggt} & 19.78 & 11.16 & 111.36 & 34.16 & 6.83 & 5.16 & 9.35 & 6.68 & 5.23 & 56.15 & 42.24 & 20.87 \\
        & Pi-Long~\cite{wang2026pi} & 19.01 & 10.82 & 170.92 & 77.59 & 5.67 & \trd{0.92} & 6.13 & \trd{4.82} & 3.24 & 36.84 & 20.27 & 23.82 \\
        & DA3-Streaming~\cite{lin2026depth} & \trd{12.72} & 9.34 & 83.64 & 37.84 & 4.32 & 3.92 & 5.87 & \snd{3.73} & 4.83 & \trd{30.82} & \trd{10.93} & 15.57 \\
        & \textbf{GRF-Recon (Ours)} & \fst{7.18} & \fst{4.35} & 76.52 & \trd{16.50} & \snd{1.65} & \snd{0.85} & \fst{3.25} & \fst{2.35} & \snd{1.64} & \snd{26.40} & \fst{7.25} & \snd{6.55} \\
        \bottomrule
    \end{tabular}
    }
\end{table*}

\subsection{Camera Tracking on Long Sequences}

Camera pose estimation is evaluated on KITTI~\cite{geiger2012we} and the Waymo Open Dataset~\cite{sun2020scalability}. All compared methods operate under a strictly monocular setting, producing scale-ambiguous trajectories. Prior to evaluation, each predicted trajectory undergoes a similarity transformation to align with the ground truth. 

As reported in Tab.~\ref{tab:kitti_tracking_heatmap_updated} and Tab.~\ref{tab:waymo_tracking_heatmap}, existing approaches struggle to maintain stability over extended sequences. DROID-SLAM~\cite{teed2021droid} accumulates severe drift on several trajectories. MASt3R-SLAM~\cite{murai2025mast3r} frequently fails when sudden scene changes enlarge the keyframe baseline, resulting in tracking loss (\textit{TL}). Feed-forward foundation models like VGGT~\cite{wang2025vggt} and the original DA3 encounter fatal out-of-memory (\textit{OOM}) errors on long sequences.

\begin{figure}[t]
  \centering
  \includegraphics[width=1.0\linewidth]{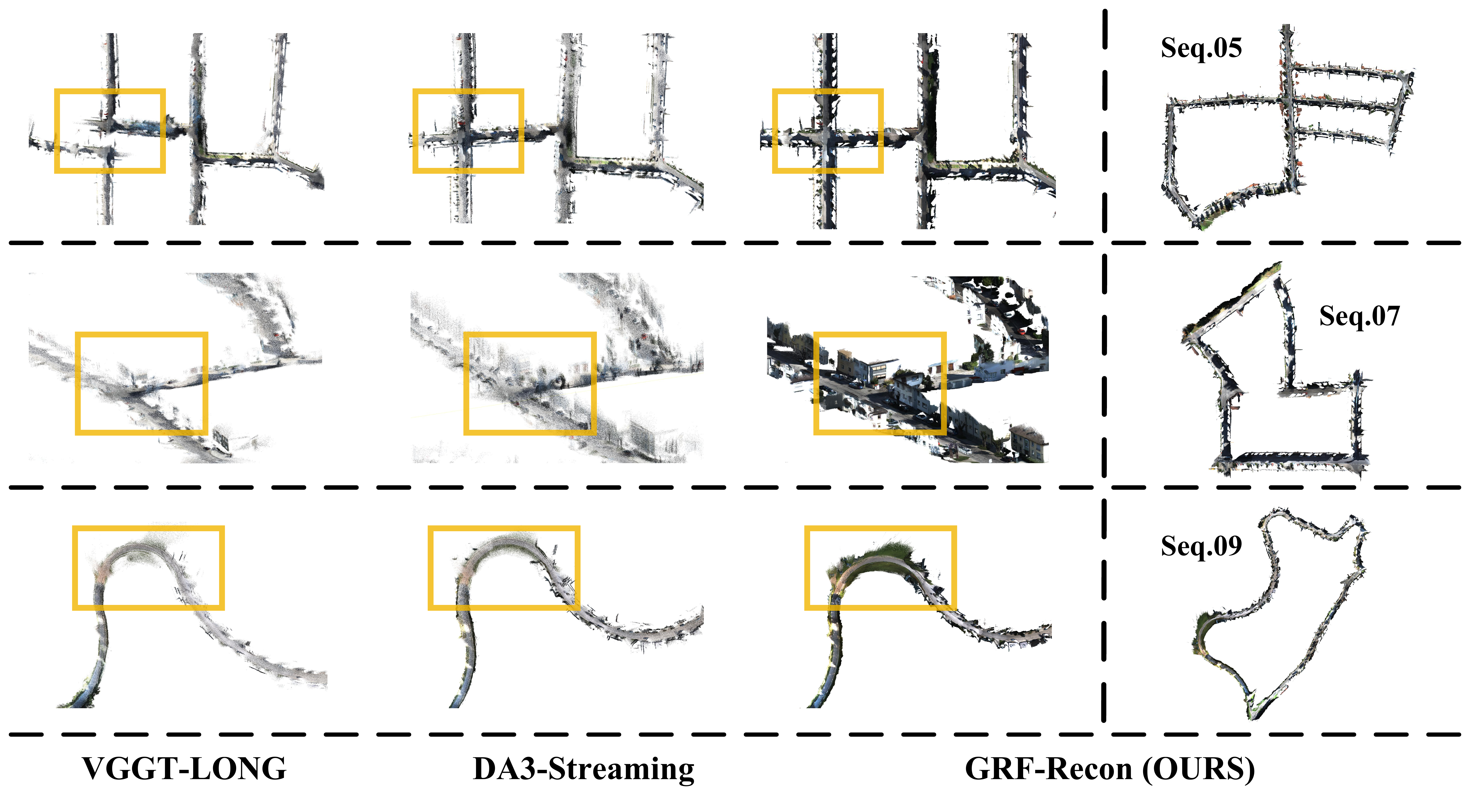} 
  \caption{\textbf{Qualitative comparison of dense local reconstruction on KITTI.}}
  \label{fig:local_detail} 
\end{figure}

As shown in Fig.~\ref{fig:local_detail}, among long-sequence feed-forward variants, DA3-Streaming maintains reasonable accuracy on most runs but exhibits clear pose discontinuities at street junctions. This limitation stems from inadequate long-range geometric consistency modeling: DA3-Streaming strictly optimizes inter-chunk pose constraints, leaving intra-chunk poses entirely reliant on feed-forward predictions. GRF-Recon overcomes this by introducing an explicit ray-field representation into the global optimization, directly encoding 3D geometry and coupling it with depth priors to enforce rigid scale constraints. This design allows our method to achieve trajectory accuracy comparable to traditional calibrated SLAM systems~\cite{campos2021orb}, entirely without requiring known camera intrinsics.

\begin{table*}[t]
    \centering
    \caption{Camera Tracking Results (ATE RMSE [m] $\downarrow$) on the Waymo Dataset~\cite{sun2020scalability}.} 
    \label{tab:waymo_tracking_heatmap}
    \resizebox{\textwidth}{!}{
    \begin{tabular}{l | c | ccccccccc}
        \toprule
        \textbf{Method} & \textbf{Avg.} & 
        \textbf{\scriptsize Seq.163} & 
        \textbf{\scriptsize Seq.183} & 
        \textbf{\scriptsize Seq.315} & 
        \textbf{\scriptsize Seq.346} & 
        \textbf{\scriptsize Seq.371} & 
        \textbf{\scriptsize Seq.405} & 
        \textbf{\scriptsize Seq.460} & 
        \textbf{\scriptsize Seq.520} & 
        \textbf{\scriptsize Seq.610} \\
        \midrule
        
        \cg{\textit{seq. frames}} & \cg{198} & \cg{198} & \cg{199} & \cg{199} & \cg{199} & \cg{196} & \cg{199} & \cg{198} & \cg{199} & \cg{198} \\
        \cg{\textit{seq. length (m)}} & \cg{172.53} & \cg{159.96} & \cg{42.30} & \cg{165.15} & \cg{351.21} & \cg{272.66} & \cg{85.74} & \cg{265.91} & \cg{134.55} & \cg{62.74} \\
        \cg{\textit{seq. speed (m/f)}} & \cg{0.87} & \cg{0.81} & \cg{0.21} & \cg{0.83} & \cg{1.77} & \cg{1.39} & \cg{0.43} & \cg{1.34} & \cg{0.68} & \cg{0.32} \\
        
        \midrule
        
        % Methods 
        DROID-SLAM~\cite{teed2021droid} & 5.43 & 3.82 & \fst{0.31} & \snd{0.47} & 8.82 & 9.45 & 7.82 & 4.25 & 13.65 & \fst{0.28} \\
        MASt3R-SLAM~\cite{murai2025mast3r} & 5.01 & 4.61 & \snd{0.59} & 1.89 & 12.85 & 8.75 & \trd{1.46} & 5.54 & 8.12 & 1.25 \\
        CUT3R~\cite{wang2025continuous} & 9.99 & 8.95 & 3.92 & 5.92 & 24.52 & 13.45 & 7.42 & 13.55 & 8.85 & 3.31 \\
        VGGT-Long~\cite{deng2026vggt} & \trd{2.03} & \trd{1.83} & 2.72 & 0.59 & \trd{3.55} & \trd{3.43} & 1.49 & \trd{1.59} & \trd{2.62} & 0.47 \\
        DA3-Streaming~\cite{lin2026depth} & \snd{1.76} & \snd{1.72} & 1.86 & \trd{0.55} & \snd{3.25} & \snd{2.92} & \snd{1.39} & \snd{1.47} & \snd{2.22} & \trd{0.42} \\
        \textbf{GRF-Recon (Ours)} & \fst{1.39} & \fst{1.49} & \trd{1.55} & \fst{0.45} & \fst{2.52} & \fst{2.15} & \fst{1.15} & \fst{1.12} & \fst{1.78} & \snd{0.29} \\
        
        \bottomrule
    \end{tabular}
    }
\end{table*}

\begin{figure}[t]
  \centering
  \includegraphics[width=1.0\linewidth]{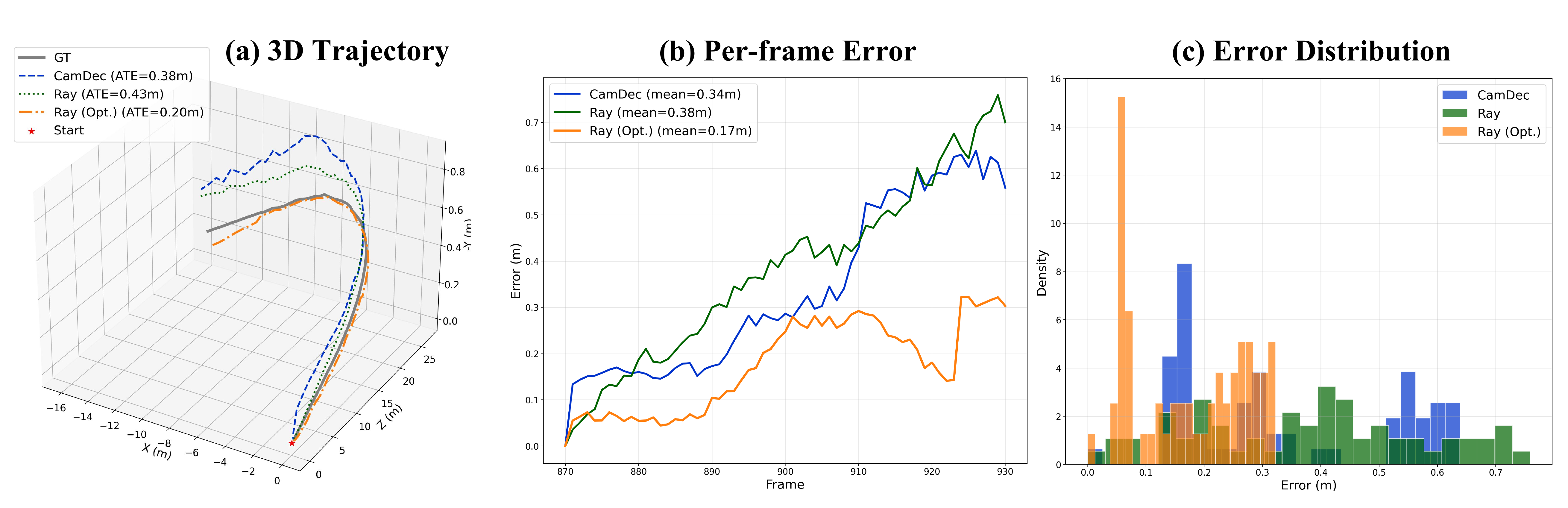} 
   \caption{\textbf{Evaluation of intra-chunk pose estimation on KITTI sequence 07.} Under loop-free conditions, our proposed ray optimization method (Ray (Opt.)) significantly mitigates local drift and improves intra-chunk pose accuracy compared to the DA3 camera decoder (CamDec) and raw ray predictions (Ray).}
  \label{fig:RAY}
\end{figure}

\begin{table}[t]
    \centering
    \setlength{\tabcolsep}{3.5pt} 
    \caption{\textbf{Runtime of GRF-Recon on Key KITTI~\cite{geiger2012we} Sequences.} End-to-end runtime excludes model loading and disk I/O. Our parallel pipeline masks the overhead of chunk alignment and ray matching behind feed-forward inference, while a sparse C++ backend minimizes global optimization cost for long sequences.}
    \label{tab:runtime_analysis}
    \resizebox{\textwidth}{!}{ 
    \begin{tabular}{lc | ccc | ccc} 
        \toprule
        \textbf{Seq.} & \textbf{Frames} & 
        \multicolumn{3}{c|}{\textbf{Frontend Pipeline}} & 
        \multicolumn{3}{c}{\textbf{Backend \& Total}} \\  
        \cmidrule(lr){3-5} \cmidrule(l){6-8}              
        
        \multicolumn{2}{c|}{\textcolor{gray}{\textit{\footnotesize Chunk Size = 60}}} & 
        \makecell{VPR Model\\(s / chunk)} & \makecell{Ray Match\\(s / keyframe)} & \makecell{Chunk Align\\(s / chunk)} & \makecell{Loop Det.\\(ms / q)} & \makecell{LM Opt.\\(s / seq)} & \makecell{Total Time} \\
        \midrule
        
        KITTI 00 & 4542 & 8.1 & 0.16 & 0.142 & 32.5 & 10.23 & 11min 45s \\
        KITTI 07 & 1101 & 8.3 & 0.17 & 0.164 & 30.7 & 2.72 & 2min 41s \\
        KITTI 09 & 1591 & 8.4 & 0.16 & 0.157 & 27.8 & 4.08 & 4min 03s \\
        \bottomrule
    \end{tabular}
    }
\end{table}

\begin{table*}[t]
    \centering
    \caption{\textbf{3D Reconstruction on Waymo Open Dataset~\cite{sun2020scalability}.} Performance is evaluated using three geometric metrics: Accuracy $\downarrow$, Completeness $\downarrow$, and Chamfer Distance $\downarrow$. All methods operate under the uncalibrated monocular setting. The best result for each metric is highlighted with \textbf{\underline{bold and underline}}. }
    \label{tab:waymo_recon_metrics}
    \resizebox{\textwidth}{!}{
    \begin{tabular}{l | l | ccccccccc}
        \toprule
        \textbf{Method} & \textbf{Metric} & 
        \textbf{\scriptsize Seq.163} & 
        \textbf{\scriptsize Seq.183} & 
        \textbf{\scriptsize Seq.315} & 
        \textbf{\scriptsize Seq.346} & 
        \textbf{\scriptsize Seq.371} & 
        \textbf{\scriptsize Seq.405} & 
        \textbf{\scriptsize Seq.460} & 
        \textbf{\scriptsize Seq.520} & 
        \textbf{\scriptsize Seq.610} \\
        \midrule
        
        % MASt3R-SLAM
        \multirow{3}{*}{MASt3R-SLAM~\cite{murai2025mast3r}} 
        & Accuracy $\downarrow$ & 3.420 & 3.250 & 4.050 & 4.950 & 4.850 & 1.250 & 5.020 & 6.850 & 2.850 \\
        & Completeness $\downarrow$ & \best{1.850} & 3.550 & 2.220 & 3.250 & 2.920 & 3.150 & 2.180 & 4.820 & 7.150 \\
        & Chamfer $\downarrow$ & 2.650 & 3.320 & 3.150 & 4.150 & 3.880 & 2.220 & 3.550 & 5.850 & 4.950 \\
        \midrule
        
        % VGGT-Long
        \multirow{3}{*}{VGGT-Long~\cite{deng2026vggt}} 
        & Accuracy $\downarrow$ & 1.120 & 0.420 & 1.020 & 1.820 & 2.750 & 0.740 & 0.850 & 1.480 & 1.350 \\
        & Completeness $\downarrow$ & 2.950 & 3.680 & 1.880 & 3.550 & 3.050 & 3.480 & 2.010 & 5.050 & 2.220 \\
        & Chamfer $\downarrow$ & 2.080 & 2.050 & 1.450 & 2.680 & 2.920 & 2.120 & \best{1.420} & 3.250 & 1.780 \\
        \midrule
        
        % DA3-Streaming
        \multirow{3}{*}{DA3-Streaming~\cite{lin2026depth}} 
        & Accuracy $\downarrow$ & 1.050 & \best{0.380} & 0.960 & 1.680 & 2.250 & 0.710 & 0.820 & 1.360 & 1.180 \\
        & Completeness $\downarrow$ & 2.680 & 3.350 & 1.680 & 3.150 & 2.780 & 3.050 & 1.820 & 4.450 & 2.020 \\
        & Chamfer $\downarrow$ & 1.820 & 1.900 & 1.350 & 2.450 & 2.650 & 1.920 & 1.580 & 2.980 & 1.580 \\
        \midrule
        
        % GRF-Recon (Ours)
        \multirow{3}{*}{\textbf{GRF-Recon (Ours)}} 
        & Accuracy $\downarrow$ & \best{0.910} & 0.410 & \best{0.820} & \best{1.350} & \best{1.980} & \best{0.620} & \best{0.740} & \best{1.150} & \best{1.020} \\
        & Completeness $\downarrow$ & 1.980 & \best{3.050} & \best{1.450} & \best{2.850} & \best{2.450} & \best{2.620} & \best{1.600} & \best{4.150} & \best{1.800} \\
        & Chamfer $\downarrow$ & \best{1.560} & \best{1.680} & \best{1.250} & \best{2.120} & \best{2.320} & \best{1.680} & 1.480 & \best{2.550} & \best{1.350} \\
        
        \bottomrule
    \end{tabular}
    }
\end{table*}

\begin{table}[t]
    \centering
    \caption{\textbf{Ablation study of GRF-Recon} (ATE RMSE [m] $\downarrow$). The entry \textit{(0,0,0)} under the Ray Comp. column indicates that the ray-matching module is completely removed. w/o Opt. disables both loop closure and ray matching, and the trajectory is obtained by directly concatenating sequential camera poses. The Baseline uses the default configuration with $C/O=60/10$.}
    \label{tab:ablation_final_professional}
    \resizebox{\textwidth}{!}{ 
    \begin{tabular}{l ccc cccc cc ccc} 
        \toprule
        \multirow{2}{*}{\textbf{Dataset}} & \multirow{2}{*}{\makecell{\textbf{Baseline}\\\textbf{(Full)}}} & \multirow{2}{*}{\makecell{\textbf{w/o}\\\textbf{LoRA}}} & \multirow{2}{*}{\makecell{\textbf{Sparse}\\\textbf{KF}}} & \multicolumn{4}{c}{\textbf{Ray Comp. $(w_r, w_g, w_p)$}} & \multirow{2}{*}{\makecell{\textbf{w/o}\\\textbf{Loop}}} & \multirow{2}{*}{\makecell{\textbf{w/o}\\\textbf{Opt.}}} & \multicolumn{3}{c}{\textbf{Config ($C/O$)}} \\
        \cmidrule(lr){5-8} \cmidrule(lr){11-13} 
        & & & & \textcolor{gray}{\textit{(1,1,0)}} & \textcolor{gray}{\textit{(1,0,1)}} & \textcolor{gray}{\textit{(0,1,1)}} & \textcolor{gray}{\textit{(0,0,0)}} & & & \textcolor{gray}{\textit{30/10}} & \textcolor{gray}{\textit{60/10}} & \textcolor{gray}{\textit{60/20}} \\
        \midrule
        KITTI 00 & \textbf{4.35} & 6.80 & 5.45 & 5.62 & 6.95 & 6.42 & 7.35 & 8.12 & 12.35 & 4.41 & 4.35 & \underline{\textbf{4.32}} \\
        KITTI 07 & \textbf{1.64} & 2.65 & 2.15 & 2.45 & 2.92 & 3.12 & 4.85 & 5.85 & 6.65 & \underline{\textbf{1.61}} & 1.64 & 1.66 \\
        \midrule
        Seq. 346 & \textbf{2.52} & 3.10 & 3.15 & 3.52 & 3.25 & 3.55 & 6.60 & 6.85 & 7.40 & 2.58 & 2.52 & \underline{\textbf{2.49}} \\
        Seq. 371 & \textbf{2.15} & 2.80 & 2.74 & 3.25 & 3.85 & 3.15 & 4.25 & 6.10 & 8.20 & 2.18 & \underline{\textbf{2.15}} & 2.17 \\
        \bottomrule
    \end{tabular}
    }
\end{table}

\subsection{Optimization and Reconstruction Quality Analysis}

We analyze system efficiency by decomposing the runtime of the frontend pipeline and backend optimization (Tab.~\ref{tab:runtime_analysis}). A lightweight trajectory-based chunking strategy replaces dense point cloud registration. 

By exploiting pipeline parallelism, the computational cost of inter-chunk ray matching ($\approx 0.16$\,s per keyframe) and chunk alignment ($\approx 0.15$\,s per chunk) is effectively hidden by the dominant feed-forward inference ($\approx 8$\,s per chunk), introducing negligible additional latency to the system. 

In the backend, our system filters noise to retain only high-confidence sparse ray constraints, resulting in a highly sparse factor graph. Even for sequences exceeding a thousand frames, the full Levenberg–Marquardt optimization converges within approximately 10\,seconds. This synergy between lightweight frontend alignment and sparse backend refinement achieves an optimal balance between trajectory accuracy and computational efficiency under a strict 24\,GB memory budget. As illustrated in Fig.~\ref{fig:RAY}, we compare the raw ray decoding from DA3~\cite{lin2026depth} with our refined ray estimation under loop-free conditions. The results demonstrate that frontend optimization effectively regularizes the trajectory and mitigates accumulated drift, confirming that high-quality ray matching inherently provides strong geometric constraints for monocular reconstruction.

Tab.~\ref{tab:waymo_recon_metrics} reports the 3D reconstruction results on Waymo~\cite{sun2020scalability}. Operating strictly under the uncalibrated monocular setting, GRF-Recon consistently achieves Accuracy, Completeness, and Chamfer Distance across most sequences. The global constraints imposed by our ray-field representation actively mitigate point cloud layering and ghosting artifacts over long trajectories, yielding structurally accurate and geometrically consistent scene reconstructions.

\subsection{Ablation Study and Applications}
\label{subsec:ablation}

Tab.~\ref{tab:ablation_final_professional} details our ablation study, quantifying the contribution of each core module. Front-end depth fine-tuning (w/o LoRA) and the hybrid-weight ray-matching module prove essential. Completely removing the ray matching module (\textit{(0,0,0)}) or disabling the global backend optimization (w/o Opt.) forces the system to degrade into pure sequential camera concatenation, triggering a drastic increase in ATE RMSE. This confirms that both components are indispensable for suppressing drift and preserving geometric consistency.

The spatio-temporal configuration ablation ($C/O$) demonstrates that, equipped with precise geometric constraints and a robust backend optimizer, the system exhibits robustness to varying chunk sizes and overlap ratios. To balance memory footprint and computation speed, we select $C/O = 60/10$ as the default baseline, securing a practical trade-off between performance and resource consumption.

\begin{figure}[t]
  \centering
  \includegraphics[width=1.0\linewidth]{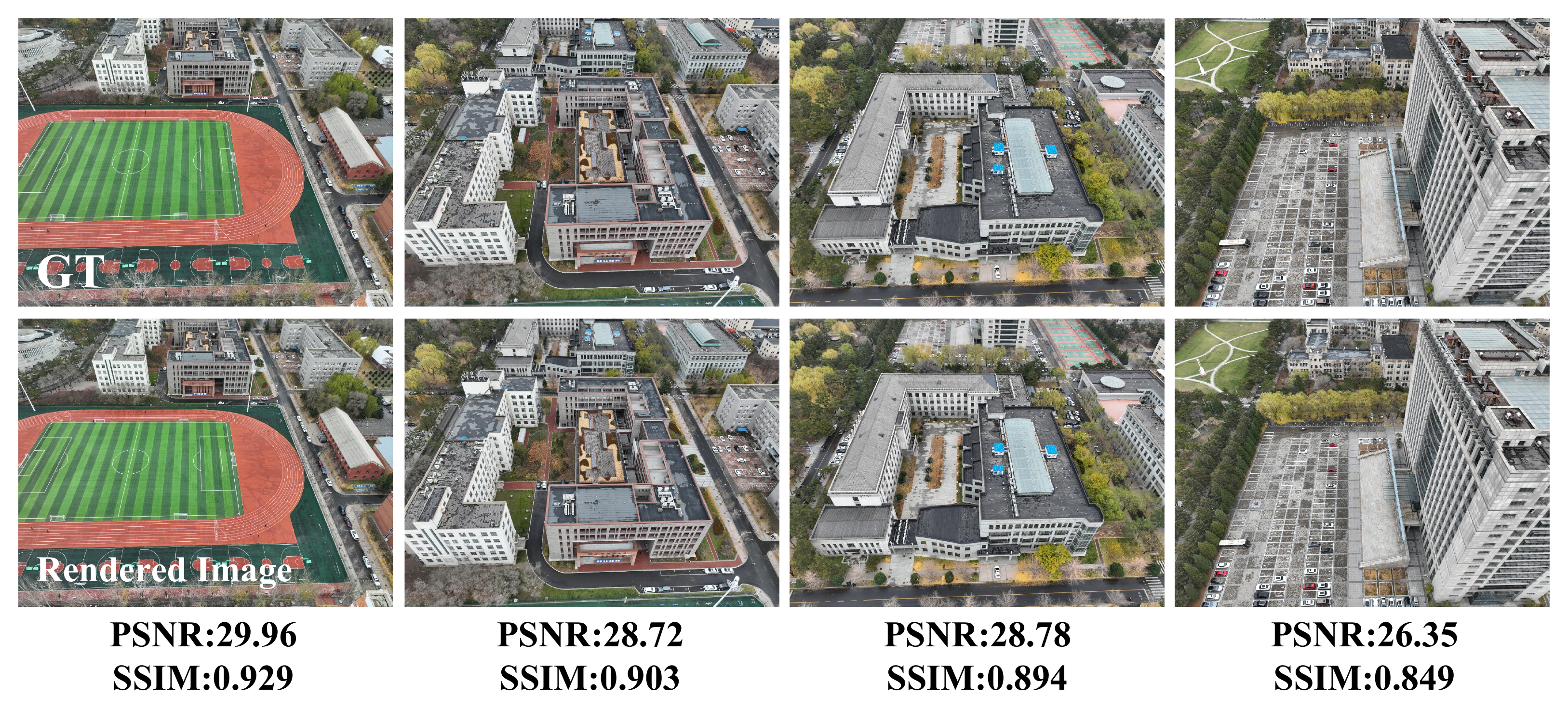} 
  \caption{\textbf{Novel view synthesis on large-scale uncalibrated UAV sequences.} We use the point clouds reconstructed by GRF-Recon to initialize 3D Gaussian Splatting, eliminating the need for a separate COLMAP-based structure-from-motion stage. With this geometry-aware initialization, 3DGS converges within approximately 1,000 iterations while maintaining accurate structural consistency. }
  \label{fig:3DGS}
\end{figure}

We deploy GRF-Recon for UAV-based urban building reconstruction (Fig.~\ref{fig:3DGS}). Operating in GPS-denied environments, the system achieves robust trajectory closure and generates high-fidelity dense point clouds. Utilizing these point clouds to initialize 3D Gaussian Splatting~\cite{kerbl20233d} accelerates convergence compared to traditional SfM-based initialization (e.g., COLMAP~\cite{schonberger2016structure}) and can effectively mitigate artifacts in edges and fine structures, demonstrating its considerable potential for downstream dense neural SLAM~\cite{chen2025vss} and rendering tasks.

\section{Conclusion}
\label{sec:conclusion}

We present GRF-Recon, a scalable framework for uncalibrated monocular 3D reconstruction over long sequences. By distilling high-fidelity geometric priors through re-normalized LoRA adaptation, we mitigate local depth degradation. To address long-term drift under strict memory constraints, we introduce a hybrid-weight sparse ray-field optimization. Experiments on autonomous driving datasets show that GRF-Recon outperforms existing feed-forward models, achieving trajectory accuracy that surpasses traditional calibrated SLAM systems in selected scenarios. The method offers a practical, geometry-aware initialization for downstream applications, including UAV-based urban mapping, virtual reality, and rapid digital twin generation via 3D Gaussian Splatting. Although robust in rigid environments, it remains challenged by highly dynamic objects. Future work will focus on improving reconstruction accuracy and real-time consistency in complex scenes.

\bibliographystyle{splncs04}
\bibliography{main}

@String(CVPR  = {IEEE Conf. Comput. Vis. Pattern Recog.})

@String(ICCV  = {Int. Conf. Comput. Vis.})

@String(ECCV  = {Eur. Conf. Comput. Vis.})

@String(NeurIPS = {Adv. Neural Inform. Process. Syst.})

@String(ICLR  = {Int. Conf. Learn. Represent.})

@String(CVPRW = {IEEE Conf. Comput. Vis. Pattern Recog. Worksh.})

@String(TMLR  = {Trans. Mach. Learn Res.})

@String(ICRA  = {IEEE Int. Conf. Robot. Autom.})

@inproceedings{
lin2026depth,
title={{Depth Anything 3}: Recovering the Visual Space from Any Views},
author={Haotong Lin and Sili Chen and Jun Hao Liew and Donny Y. Chen and Zhenyu Li and Yang Zhao and Sida Peng and Hengkai Guo and Xiaowei Zhou and Guang Shi and Jiashi Feng and Bingyi Kang},
booktitle= ICLR,
year={2026}
}

@inproceedings{
ren2025finr,
title={{Fin3R}: Fine-tuning Feed-forward {3D} Reconstruction Models via Monocular Knowledge Distillation},
author={Weining Ren and Hongjun Wang and Xiao Tan and Kai Han},
booktitle= NeurIPS,
year={2025}
}

@inproceedings{wang2025moge,
  title={{Moge}: Unlocking accurate monocular geometry estimation for open-domain images with optimal training supervision},
  author={Wang, Ruicheng and Xu, Sicheng and Dai, Cassie and Xiang, Jianfeng and Deng, Yu and Tong, Xin and Yang, Jiaolong},
  booktitle= CVPR,
  pages={5261--5271},
  year={2025}
}

@article{campos2021orb,
  author       = {Carlos Campos and
                  Richard Elvira and
                  Juan J. G{\'{o}}mez Rodr{\'{\i}}guez and
                  Jos{\'{e}} M. M. Montiel and
                  Juan D. Tard{\'{o}}s},
  title        = {{ORB-SLAM3}: An Accurate Open-Source Library for Visual, Visual-Inertial,
                  and Multimap {SLAM}},
  journal      = {{IEEE} Trans. Robotics},
  volume       = {37},
  number       = {6},
  pages        = {1874--1890},
  year         = {2021}
}

@inproceedings{teed2021droid,
  author       = {Zachary Teed and
                  Jia Deng},
  title        = {{DROID-SLAM}: Deep Visual {SLAM} for Monocular, Stereo, and {RGB-D}
                  Cameras},
  booktitle    =  NeurIPS,
  pages        = {16558--16569},
  year         = {2021}
}

@inproceedings{teed2023deep,
  author       = {Zachary Teed and
                  Lahav Lipson and
                  Jia Deng},
  title        = {Deep Patch Visual Odometry},
  booktitle    = NeurIPS,
  year         = {2023}
}

@inproceedings{lipson2024deep,
  title={Deep patch visual slam},
  author={Lipson, Lahav and Teed, Zachary and Deng, Jia},
  booktitle= ECCV,
  pages={424--440},
  year={2024}
}

@inproceedings{wang2025vggt,
  title={{Vggt}: Visual geometry grounded transformer},
  author={Wang, Jianyuan and Chen, Minghao and Karaev, Nikita and Vedaldi, Andrea and Rupprecht, Christian and Novotny, David},
  booktitle= CVPR,
  pages={5294--5306},
  year={2025}
}

@inproceedings{deng2026vggt,
  title={{VGGT-Long}: Chunk it, Loop it, Align it -- Pushing {VGGT}'s Limits on Kilometer-scale Long {RGB} Sequences},
  author={Deng, Kai and Ti, Zexin and Xu, Jiawei and Yang, Jian and Xie, Jin},
  booktitle = ICRA,
  year={2026}
}

@inproceedings{leroy2024grounding,
  title={Grounding Image Matching in {3D} with {MASt3R}},
  author={Leroy, Vincent and Cabon, Yohann and Revaud, J{\'e}r{\^o}me},
  booktitle= ECCV,
  pages={71--91},
  year={2024},
}

@inproceedings{murai2025mast3r,
  title={{Mast3r-slam}: Real-time dense slam with {3D} reconstruction priors},
  author={Murai, Riku and Dexheimer, Eric and Davison, Andrew J},
  booktitle= CVPR,
  pages={16695--16705},
  year={2025}
}

@inproceedings{wang2024dust3r,
  title={{Dust3r}: Geometric {3D} vision made easy},
  author={Wang, Shuzhe and Leroy, Vincent and Cabon, Yohann and Chidlovskii, Boris and Revaud, Jerome},
  booktitle= CVPR,
  pages={20697--20709},
  year={2024}
}

@inproceedings{wang2025continuous,
  title={Continuous {3D} perception model with persistent state},
  author={Wang, Qianqian and Zhang, Yifei and Holynski, Aleksander and Efros, Alexei A and Kanazawa, Angjoo},
  booktitle= CVPR,
  pages={10510--10522},
  year={2025}
}

@inproceedings{wang2026pi,
  title={{$\pi^3$}: Permutation-Equivariant Visual Geometry Learning},
  author={Yifan Wang and Jianjun Zhou and Haoyi Zhu and Wenzheng Chang and Yang Zhou and Zizun Li and Junyi Chen and Jiangmiao Pang and Chunhua Shen and Tong He},
  booktitle= ICLR,
  year={2026}
}

@inproceedings{schonberger2016structure,
  title={Structure-from-motion revisited},
  author={Schonberger, Johannes L and Frahm, Jan-Michael},
  booktitle=CVPR,
  pages={4104--4113},
  year={2016}
}

@inproceedings{detone2018superpoint,
  title={{Superpoint}: Self-supervised interest point detection and description},
  author={DeTone, Daniel and Malisiewicz, Tomasz and Rabinovich, Andrew},
  booktitle= CVPRW,
  pages={224--236},
  year={2018}
}

@inproceedings{sarlin2020superglue,
  title={{Superglue}: Learning feature matching with graph neural networks},
  author={Sarlin, Paul-Edouard and DeTone, Daniel and Malisiewicz, Tomasz and Rabinovich, Andrew},
  booktitle= CVPR,
  pages={4937-4946},
  year={2020}
}

@book{hartley2003multiple,
  title={Multiple view geometry in computer vision},
  author={Hartley, Richard and Zisserman, Andrew},
  year={2003},
  publisher={Cambridge university press}
}

@inproceedings{wang2024vggsfm,
  title={{Vggsfm}: Visual geometry grounded deep structure from motion},
  author={Wang, Jianyuan and Karaev, Nikita and Rupprecht, Christian and Novotny, David},
  booktitle=CVPR,
  pages={21686--21697},
  year={2024}
}

@inproceedings{vaswani2017attention,
  author       = {Ashish Vaswani and
                  Noam Shazeer and
                  Niki Parmar and
                  Jakob Uszkoreit and
                  Llion Jones and
                  Aidan N. Gomez and
                  Lukasz Kaiser and
                  Illia Polosukhin},
  title        = {Attention is All you Need},
  booktitle    = NeurIPS,
  pages        = {5998--6008},
  year         = {2017}
}

@inproceedings{wang20253dspann3r,
  title={{3D} reconstruction with spatial memory},
  author={Wang, Hengyi and Agapito, Lourdes},
  booktitle={2025 International Conference on {3D} Vision (3DV)},
  pages={78--89},
  year={2025}
}

@inproceedings{chen2025long3r,
  title={{Long3r}: Long sequence streaming {3D} reconstruction},
  author={Chen, Zhuoguang and Qin, Minghui and Yuan, Tianyuan and Liu, Zhe and Zhao, Hang},
  booktitle=ICCV,
  pages={5273--5284},
  year={2025}
}

@inproceedings{cabon2025must3r,
  title={{Must3r}: Multi-view network for stereo {3D} reconstruction},
  author={Cabon, Yohann and Stoffl, Lucas and Antsfeld, Leonid and Csurka, Gabriela and Chidlovskii, Boris and Revaud, Jerome and Leroy, Vincent},
  booktitle= CVPR,
  pages={1050--1060},
  year={2025}
}

@inproceedings{yang2025fast3r,
  title={{Fast3r}: Towards {3D} reconstruction of 1000+ images in one forward pass},
  author={Yang, Jianing and Sax, Alexander and Liang, Kevin J and Henaff, Mikael and Tang, Hao and Cao, Ang and Chai, Joyce and Meier, Franziska and Feiszli, Matt},
  booktitle= CVPR,
  pages={21924--21935},
  year={2025}
}

@inproceedings{rao2021dynamicvit,
  title= {{DynamicViT}: Efficient Vision Transformers with Dynamic Token Sparsification},
  author= {Yongming Rao and Wenliang Zhao and Benlin Liu and Jiwen Lu and Jie Zhou and Cho{-}Jui Hsieh},
  booktitle= NeurIPS,
  pages= {13937--13949},
  year= {2021}
}

@inproceedings{bolya2022token,
  title={{Token Merging}: Your ViT But Faster},
  author={Daniel Bolya and Cheng-Yang Fu and Xiaoliang Dai and Peizhao Zhang and Christoph Feichtenhofer and Judy Hoffman},
  booktitle=ICLR,
  year={2023}
}

@inproceedings{shen2026fastvggt,
  title={Fast{VGGT}: Fast Visual Geometry Transformer},
  author={You Shen and Zhipeng Zhang and Yansong Qu and Xiawu Zheng and Jiayi Ji and Shengchuan Zhang and Liujuan Cao},
  booktitle=ICLR,
  year={2026}
}

@inproceedings{hu2022lora,
  title        = {{LoRA}: Low-Rank Adaptation of Large Language Models},
  author       = {Edward J. Hu and Yelong Shen and Phillip Wallis and Zeyuan Allen{-}Zhu and Yuanzhi Li and Shean Wang and Lu Wang and Weizhu Chen},
  booktitle    = ICLR,
  year         = {2022}
}

@article{
oquab2024dinov2,
title={{DINO}v2: Learning Robust Visual Features without Supervision},
author={Maxime Oquab and Timoth{\'e}e Darcet and Th{\'e}o Moutakanni and Huy V. Vo and Marc Szafraniec and Vasil Khalidov and Pierre Fernandez and Daniel HAZIZA and Francisco Massa and Alaaeldin El-Nouby and Mido Assran and Nicolas Ballas and Wojciech Galuba and Russell Howes and Po-Yao Huang and Shang-Wen Li and Ishan Misra and Michael Rabbat and Vasu Sharma and Gabriel Synnaeve and Hu Xu and Herve Jegou and Julien Mairal and Patrick Labatut and Armand Joulin and Piotr Bojanowski},
journal=TMLR,
issn={2835-8856},
year={2024},
note={Featured Certification}
}

@article{kerbl20233d,
  author       = {Bernhard Kerbl and
                  Georgios Kopanas and
                  Thomas Leimk{\"{u}}hler and
                  George Drettakis},
  title        = {{3D} Gaussian Splatting for Real-Time Radiance Field Rendering},
  journal      = {{ACM} Trans. Graph.},
  volume       = {42},
  number       = {4},
  pages        = {139:1--139:14},
  year         = {2023}
}

@inproceedings{geiger2012we,
  title={Are we ready for autonomous driving? the kitti vision benchmark suite},
  author={Geiger, Andreas and Lenz, Philip and Urtasun, Raquel},
  booktitle=CVPR,
  pages={3354--3361},
  year={2012}
}

@inproceedings{sun2020scalability,
  author= {Pei Sun and Henrik Kretzschmar and Xerxes Dotiwalla and Aurelien Chouard and Vijaysai Patnaik and Paul Tsui and James Guo and Yin Zhou and Yuning Chai and Benjamin Caine and Vijay Vasudevan and Wei Han and Jiquan Ngiam and Hang Zhao and Aleksei Timofeev and Scott Ettinger and Maxim Krivokon and Amy Gao and Aditya Joshi and Yu Zhang and Jonathon Shlens and Zhifeng Chen and Dragomir Anguelov},
  title        = {Scalability in Perception for Autonomous Driving: Waymo Open Dataset},
  booktitle    = CVPR,
  pages        = {2443--2451},
  year         = {2020}
}

@inproceedings{schops2017multi,
  title={A multi-view stereo benchmark with high-resolution images and multi-camera videos},
  author={Schops, Thomas and Schonberger, Johannes L and Galliani, Silvano and Sattler, Torsten and Schindler, Konrad and Pollefeys, Marc and Geiger, Andreas},
  booktitle=CVPR,
  pages={2538--2547},
  year={2017}
}

@article{baker2011database,
  author= {Simon Baker and Daniel Scharstein and J. P. Lewis and Stefan Roth and Michael J. Black and Richard Szeliski},
  title        = {A Database and Evaluation Methodology for Optical Flow},
  journal      = {Int. J. Comput. Vis.},
  volume       = {92},
  number       = {1},
  pages        = {1--31},
  year         = {2011}
}

@misc{vasiljevic2019diode,
      title={{DIODE}: A Dense Indoor and Outdoor DEpth Dataset}, 
      author={Igor Vasiljevic and Nick Kolkin and Shanyi Zhang and Ruotian Luo and Haochen Wang and Falcon Z. Dai and Andrea F. Daniele and Mohammadreza Mostajabi and Steven Basart and Matthew R. Walter and Gregory Shakhnarovich},
      year={2019},
      eprint={1908.00463},
      archivePrefix={arXiv},
      primaryClass={cs.CV},
      url={https://arxiv.org/abs/1908.00463},
      note = {{Accessed}: 29 June 2026}
}

@inproceedings{tyszkiewicz2020disk,
  author       = {Michal J. Tyszkiewicz and Pascal Fua and Eduard Trulls},
  title        = {{DISK}: Learning local features with policy gradient},
  booktitle    = NeurIPS,
  year         = {2020}
}

@inproceedings{lindenberger2023lightglue,
  title={{Lightglue}: Local feature matching at light speed},
  author={Lindenberger, Philipp and Sarlin, Paul-Edouard and Pollefeys, Marc},
  booktitle=ICCV,
  pages={17581--17592},
  year={2023}
}

@inproceedings{triggs1999bundle,
  author       = {Bill Triggs and
                  Philip F. McLauchlan and
                  Richard I. Hartley and
                  Andrew W. Fitzgibbon},
  title        = {Bundle Adjustment - {A} Modern Synthesis},
  booktitle    = {Workshop on Vision Algorithms},
  series       = {Lecture Notes in Computer Science},
  volume       = {1883},
  pages        = {298--372},
  publisher    = {Springer},
  year         = {1999}
}

@inproceedings{newcombe2011dtam,
  title={{DTAM}: Dense tracking and mapping in real-time},
  author={Newcombe, Richard A and Lovegrove, Steven J and Davison, Andrew J},
  booktitle=ICCV,
  pages={2320--2327},
  year={2011}
}

@article{czarnowski2020deepfactors,
  title={{Deepfactors}: Real-time probabilistic dense monocular slam},
  author={Czarnowski, Jan and Laidlow, Tristan and Clark, Ronald and Davison, Andrew J},
  journal={IEEE Robotics and Automation Letters},
  volume={5},
  number={2},
  pages={721--728},
  year={2020},
  publisher={IEEE}
}

@inproceedings{zhang2023go,
  title={{Go-slam}: Global optimization for consistent {3D} instant reconstruction},
  author={Zhang, Youmin and Tosi, Fabio and Mattoccia, Stefano and Poggi, Matteo},
  booktitle=ICCV,
  pages={3704--3714},
  year={2023}
}

@misc{birkl2023midas,
      title={{MiDaS v3.1} -- A Model Zoo for Robust Monocular Relative Depth Estimation}, 
      author={Birkl, Reiner and Wofk, Diana and M{\"u}ller, Matthias},
      year={2023},
      eprint={2307.14460},
      archivePrefix={arXiv},
      primaryClass={cs.CV},
      url={https://arxiv.org/abs/2307.14460},
      note = {{Accessed}: 29 June 2026}
}

@misc{sola2021micro,
      title={A micro Lie theory for state estimation in robotics}, 
      author={Joan Solà and Jeremie Deray and Dinesh Atchuthan},
      year={2021},
      eprint={1812.01537},
      archivePrefix={arXiv},
      primaryClass={cs.RO},
      url={https://arxiv.org/abs/1812.01537},
      note = {{Accessed}: 29 June 2026}
}

@inproceedings{tolias2013aggregate,
  title={To aggregate or not to aggregate: Selective match kernels for image search},
  author={Tolias, Giorgos and Avrithis, Yannis and J{\'e}gou, Herv{\'e}},
  booktitle=ICCV,
  pages={1401--1408},
  year={2013}
}

@inproceedings{chen2025vss,
  title={{VSS-SLAM}: Voxelized surfel splatting for geometally accurate SLAM},
  author={Chen, Xuanhua and Zhang, Yunzhou and Zhang, Zhiyao and Wang, Guoqing and Zhao, Bin and Wang, Xingshuo},
  booktitle=ICRA,
  pages={139--145},
  year={2025}
}
\end{document}